\documentclass[letterpaper, 10 pt, conference]{ieeeconf}  

\IEEEoverridecommandlockouts                              

\usepackage{graphicx}
\usepackage{amsmath}
\usepackage{amssymb}
\usepackage{booktabs}
\usepackage{hyperref}

\title{\LARGE \bf
Amirkabir campus dataset: Real-world challenges and scenarios of Visual Inertial Odometry (VIO) for visually impaired people
}

\author{Ali Samadzadeh$^{1}$, Mohammad Hassan Mojab$^{2}$, Heydar Soudani$^{2}$, \\Seyed Hesamoddin Mireshghollah$^{1}$, Ahmad Nickabadi
\thanks{$^{1}$Department of Computer Engineering,
Amirkabir University of Technology, Tehran, Iran {\tt\small \{a\_samad, h.mireshgholah, nickabadi\}@aut.ac.ir}}%
\thanks{$^{2}$Department of Computer Engineering, 
Sharif University of Technology, Tehran, Iran
        {\tt\small\{mhmojab, heydars\}@ce.sharif.edu}}%
}

\begin{document}

\maketitle
\thispagestyle{empty}
\pagestyle{empty}

\begin{abstract}
Visual Inertial Odometry (VIO) algorithms estimate the accurate camera trajectory by using camera and Inertial Measurement Unit (IMU) sensors. The applications of VIO span a diverse range, including augmented reality and indoor navigation. VIO algorithms hold the potential to facilitate navigation for visually impaired individuals in both indoor and outdoor settings. Nevertheless, state-of-the-art VIO algorithms encounter substantial challenges in dynamic environments, particularly in densely populated corridors. Existing VIO datasets, e.g., ADVIO, typically fail to effectively exploit these challenges. In this paper, we introduce the Amirkabir campus dataset (AUT-VI) to address the mentioned problem and improve the navigation systems. AUT-VI is a novel and super-challenging dataset with 126 diverse sequences in 17 different locations. This dataset contains dynamic objects, challenging loop-closure/map-reuse, different lighting conditions, reflections, and sudden camera movements to cover all extreme navigation scenarios. Moreover, in support of ongoing development efforts, we have released the Android application for data capture to the public. This allows fellow researchers to easily capture their customized VIO dataset variations. In addition, we evaluate state-of-the-art Visual Inertial Odometry (VIO) and Visual Odometry (VO) methods on our dataset, emphasizing the essential need for this challenging dataset.
\end{abstract}

\let\thefootnote\relax\footnotetext{The dataset and the application are open-source and free for the public at \url{https://a3dv.github.io/autvi}}

\section{Introduction}
\label{sec:intro}

Designing navigation systems for individuals with visual impairments is important for researchers and developers. A critical element of such systems is a robust localization algorithm capable of accurately determining the user's position and direction. Typically, this localization relies on data collected from the device's integrated sensors. Using GPS data has always been a solution in outdoor environments. However, there are limitations associated with relying solely on GPS, as extensively discussed in the literature.

On the other hand, the sensors used in localization methods should be inexpensive, available, and easy to use for blind people. Nowadays, smartphones are equipped with various sensors including cameras, accelerometers, and gyroscopes. Consequently, the localization algorithms that rely on the data provided by the sensors within the smartphones are a suitable solution.

\begin{figure} 
    \centering
    \includegraphics[width=\linewidth]{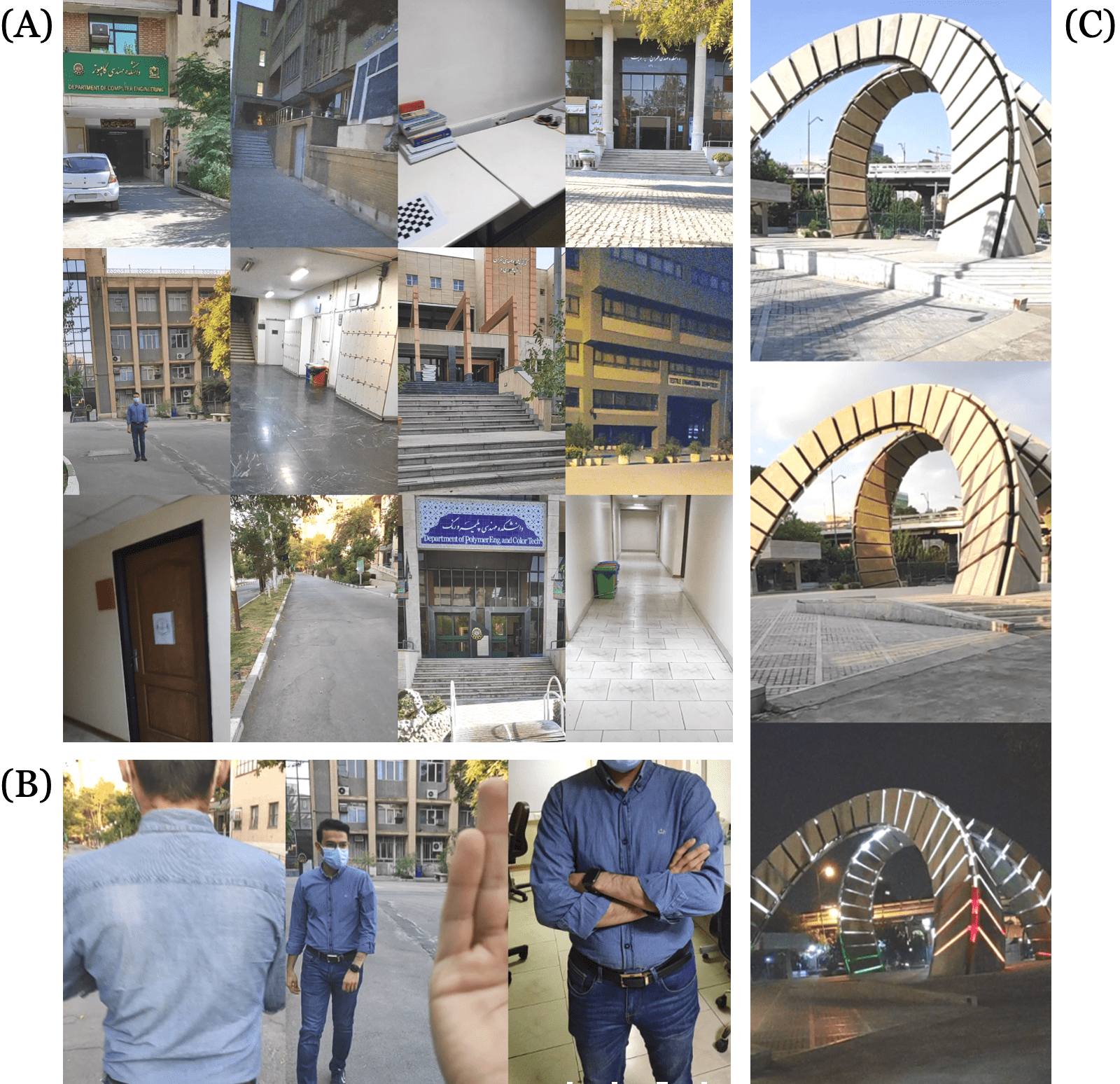}
    \caption{Example scenes from the AUT-VI dataset. A) There are multiple locations (indoors and outdoors) over the campus where the dataset is recorded. B) More than half of the sequences in the dataset are completely dynamic and challenging. C) The videos are captured at various times of the day for each location with almost the same trajectory.}
    \label{fig:all}
    \vskip -0.2in
\end{figure}

Visual Inertial Odometry (VIO) algorithms determine the location and orientation of an entity in a global and metric coordination system using a visual sensor (camera) and inertial sensor data. VIO algorithms designed to run on smartphones can rely on visual feed and inertial data to estimate the motion. A visual sensor alone cannot ensure the system's robustness and usually yields significant relative errors. Usually, Inertial Measurement Unit (IMU) data is used to reduce this error. Unlike the camera that captures environmental clues of a movement, IMU measures the internal state of the moving object that is not affected by the environment~\cite{minoda2021viode}.

The VIO algorithms must deal with various challenges to achieve accurate localization. One of the most significant challenges is dynamic objects~\cite{bescos2018dynaslam} such as humans, vehicles, and other moving objects~\cite{xiao2019dynamic}. The main issue of having dynamic objects inside the camera view is finding consistent visual features and using them to build a stable and accurate 3D point cloud and sequence of camera poses through time for further usage. 

Although there are VIO algorithms that claim to consider dynamic objects~\cite{mu2019visual, bai2020perception}, it is difficult to evaluate the performance of these algorithms in real-world dynamic situations as there is no suitable dataset for this purpose.
The second significant challenge of the VIO frameworks is loop closure and pose optimization~\cite{qin2018vins}. The loop closure's goal is to globally optimize the poses such that it reduces the long-term accumulated error and improves robustness. The loop closure is claimed to be a crucial part of the modern SLAM frameworks~\cite{campos2021orb}. Additionally, there are other challenges such as camera shaking, changing the angle of the camera along the way, and illumination change~\cite{minoda2021viode}. To the best of our knowledge, no VIO algorithm has solved all of the challenges mentioned earlier. More importantly, there is no dataset to evaluate VIO algorithms against all these challenges.

In order to evaluate the VIO algorithms in the presence of the aforementioned challenges, we propose the AUT-VI dataset, which is the most challenging dynamic VIO dataset. This benchmark helps the evaluation and improvement of navigation tools designed for visually impaired people. 
In the context of dynamic objects, the AUT-VI dataset contains different scenarios, such as dynamic objects that may entirely cover the camera view for more than a few seconds. There are also many cases where people move in the scene. In addition to people, other dynamic objects such as cars and motorcycles are also considered.
To address the challenge of robust loop-closure/map-reuse, we have designed different scenarios in both indoor and outdoor environments. There are sequences with frames at several angles in different lighting conditions (such as darkness, dimness, and lightness) and at different times of day (including morning, noon, and night) or in shaded areas.
We have captured the mentioned scenarios in several locations, including different rooms, corridors, and entrances of the Amirkabir University of Technology's departments. 
Furthermore, we have included various types of light source challenges and reflections, such as looking in a mirror or glass. In addition, we have considered different camera holding setups: standard upright holding, casual walk holding, and a scenario in which the phone is passed from one hand to another.
As far as we are concerned, no dataset covers all these challenges. Fig.~\ref{fig:all} shows some examples from the \mbox{AUT-VI} dataset.

Moreover, while some current datasets (e.g. EuRoC MAV~\cite{burri2016euroc}) are limited in the number and variety of the scenes, we have created 126 diverse sequences in 17 different places.
Furthermore, we have evaluated the state-of-the-art (SOTA) VIO algorithms against our dataset to see how they deal with the existing challenges.

Finally, one major contribution is that we have designed an application that enables the researchers to replicate or customize our dataset and even create their own datasets using their phones. 

\section{Related Works}
\label{sec:related}

\subsection{VIO datasets}

There are many VIO datasets available each containing a specific challenge. The challenges include dynamic objects, robust loop-closure/map-reuse, and lighting conditions. Even though the available datasets are suitable for evaluating the VIO algorithms in simple situations, they do not cover most of the challenges in realistic scenarios. In addition to being authentic and reflecting real-world scenarios, a VIO dataset should contain challenging environments. 
None of the available datasets encompass all of the realistic challenges, and none of them are suitable for developing visually impaired people's navigation systems. In the following, these datasets are reviewed in more detail.

The first category of odometry datasets only provides the image frames~\cite{cordts2016cityscapes, smith2009new}. In addition to only having visual data, these datasets suffer from other limitations. For example, they usually do not have rapid rotations nor do they not contain isolated low-textured indoor scenes.
TUM RGB-D~\cite{sturm2012benchmark} is another dataset in this category. This dataset has focused on evaluating RGB-D odometry and SLAM algorithms and has been extensively used by the research community. It provides 47 RGB-D sequences with ground-truth pose trajectories recorded with a motion capture system. 
However, it is limited to small environments and is not suitable for benchmarking odometry for medium and long-range navigation.
TUM MonoVO~\cite{engel2016photometrically} is another dataset in this group that contains 50 sequences in indoor and outdoor environments. It has been photometrically calibrated for exposure times, lens vignetting, and camera response functions. 

In contrast to the above pure vision datasets, some datasets contain inertial sensor data in addition to visual data. 
For instance, KITTI~\cite{geiger2013vision} and Málaga Urban~\cite{blanco2014malaga} datasets include low-frequency IMU information. The rest of the datasets are categorized and discussed in the following.

Visual Inertial (VI) datasets can be categorized according to the challenges they include.
EuRoC MAV dataset~\cite{burri2016euroc} only contains a small number of people moving in indoor scenes, and the scenes are mostly static. This dataset includes 11 short indoor sequences with a small variety of recordings in one machine hall and one lab room. Furthermore, EuRoC MAV does not include a photometric calibration. However, this dataset contains good challenges, e.g. moving at high speed and different lighting conditions.

Outdoor VI datasets usually tend to contain more dynamic objects. The TUM VI~\cite{schubert2018tum} and the Zurich Urban~\cite{majdik2017zurich} datasets contain outdoor sequences where occasionally moving vehicles and people appear. In addition, the TUM VI dataset has good optical challenges. However, it has minimal loop-closure challenges, only involving changing the angle. Also, its ground-truth is partial.
KITTI~\cite{geiger2013vision}, Urban@CRAS~\cite{gaspar2018urban}, and KAIST Urban~\cite{jeong2019complex} are outdoor datasets recorded from driving cars in urban areas. These datasets contain several moving vehicles. However, the camera's field of view is mainly occupied by static objects and as a result they are not challenging enough to be a benchmark for evaluating robustness of dynamic SLAM algorithm. Also, KAIST Urban adds some more challenges but does not have IMU and camera calibration parameters.
Another popular VI dataset is ADVIO~\cite{cortes2018advio} which includes challenges such as reflective objects and dynamic objects. However, it has limited sequences with extremely dynamic objects. Also, it does not provide accurate ground-truth.
The latest dataset of this category is VIODE~\cite{minoda2021viode}, which covers many challenges but is created in a simulated environment and is not a real-world dataset.

\renewcommand{\arraystretch}{1.2}
\begin{table*}
	\centering
	\begin{tabular}{p{5.5cm} p{.7cm} p{1.6cm} p{1.5cm} p{1.2cm} p{1.2cm} p{1.2cm} p{1.2cm}} 
		\hline
		Dataset & Year & Carrier & Dynamic. & Day/ Night & Reflect. chal. & Camera chal. & Reprod \\ [0.5ex] 
		\hline
		TUM RGB-D \cite{sturm2012benchmark} 
			& 2012
			& Handheld
			& Low\
			& No
			& No
			& Mid 
			& No \\
		KITTI \cite{geiger2013vision} 
			& 2013 
			& Car 
			& Low
			& No
			& No
			& Low
			& No \\ 
		EuRoC MAV \cite{burri2016euroc} 
			& 2016
			& UAV
			& Low
			& No
			& No
			& Low 
			& No \\ 
		Zurich Urban \cite{majdik2017zurich} 
			& 2017
			& UAV
			& Mid
			& No
			& No
			& Low 
			& No \\ 
		PennCOSYVIO \cite{pfrommer2017penncosyvio} 
			& 2017
			& Handheld 
			& Low
			& No
			& No
			& Low 
			& No \\ 
		TUM VI \cite{schubert2018tum} 
			& 2018
			& UAV
			& Low
			& No
			& No
			& Low 
			& No \\ 
		ADVIO \cite{cortes2018advio} 
			& 2018
			& Handheld 
			& High
			& No
			& Yes
			& Mid 
			& No \\ 
		Urban@CRAS \cite{gaspar2018urban} 
			& 2018
			& Car
			& Mid
			& Yes
			& No
			& Low 
			& No \\ 
		Oxford Multimotion \cite{judd2019oxford} 
			& 2019
			& Handheld
			& Mid
			& No
			& No
			& Low 
			& No \\ 
		KAIST Urban \cite{jeong2019complex} 
			& 2019
			& Car
			& Mid
			& Yes
			& No
			& Low 
			& No \\ 						
		OIVIO \cite{kasper2019benchmark} 
			& 2019
			& Handheld
			& Low
			& No
			& No
			& Mid 
			& No \\ 
		UMA-VI \cite{zuniga2020vi} 
			& 2020
			& Handheld 
			& Low
			& No
			& No
			& Mid 
			& No \\ 
		VIODE \cite{minoda2021viode} 
			& 2021
			& UAV (sim.)
			& High
			& Yes
			& No
			& Low 
			& No \\ 
		\hline
		{\bf AUT-VI (ours)}
			& 2022
			& Handheld 
			& Very High
			& Yes
			& Yes
			& High
			& Yes \\ 			
		\hline
	\end{tabular}
	\caption{An overview of related datasets.}
	\label{table:datasets}
\end{table*}

It is clear that the existing VI datasets are not general and do not cover all of the challenges in a real-world scenario. Therefore, we decided to represent the AUT-VI dataset. This dataset is much more complete than the current datasets and contains challenges of (1) dynamic objects, (2) robust loop-closure/map-reuse, (3) lighting conditions, (4) reflection, and (5) sharp movements. In addition, we put each challenge in its own isolated sequence to make the evaluation easier. We also have a special-purpose application that distinguishes our AUT-VI from the other datasets and enables the researchers to record their own data using only a smartphone.

In \ref{table:datasets}, we provide a complete summary of existing VI datasets and our AUT-VI dataset. In the dynamicity column, ``very high'' means that we have unique sequences designed to exploit possible dynamic environment scenarios. Scenarios include walking behind a person which covers more than 80\% of the camera view for at least 10 seconds, or two persons in front of the camera which cover almost all of the camera, one static and the other dynamic. Also, some scenarios are designed to include every possible dynamic sequence in a regular campus walk. By our definition, high dynamic datasets are the ones that contain at least a frame in which a person covers 80\% of the camera.

Regarding reflectiveness, there are two groups of sequences containing highly reflective objects; the library and the Polymer department. All previous datasets contain a small number of scenes with reflective objects, for comparison, ADVIO contains sequences with store glasses and mirrors but none of them cover more than 80\% of the camera view. In this dataset, we have sequences in which the camera looks into a reflective mirror completely which is highly challenging for odometry algorithms.

Furthermore, for more comprehensive quantitative comparisons, in \ref{tab:quantitative_comparisons}, we compared AUT-VI with ADVIO, a rival dataset, in terms of the duration of camera coverage for each challenge. The results show that ADVIO has only 25 minutes of dynamic scenes while our dataset contains more than 42 minutes of dynamic scenes, 7 minutes of which include dynamic objects covering the entire camera view. The total duration of scenes containing reflective objects is about 9 and 5 minutes for ADVIO and our dataset, respectively. Finally, our dataset contains more than 2 hours of camera challenge scenes, while less than one minute of the ADVIO dataset can be considered a camera challenge. 

\setlength{\tabcolsep}{16pt}
\begin{table}[t] 
    \centering
    \begin{small}
        \begin{tabular}{lrr} 
            \hline
            Challenges                & ADVIO  & AUT-VI \\
            \hline
            Camera challenge          & 49 s    & 2h25m10 s \\
            Reflective objects        & 9m10 s  & 5m43 s   \\
            Dynamic objects           & 25m42 s & 42m5 s \\
            Full cover                & 6 s     & 7m9 s  \\
            \hline
        \end{tabular}
    \end{small}
    \caption{Quantitative comparisons of AUT-VI with ADVIO, in terms of duration of camera coverage for each challenge.}
    \label{tab:quantitative_comparisons}
\end{table}

\section{AUT-VI dataset}
\label{sec:dataset}
In this section, the details of the recording rig, sequences, statistics of the dataset, and calibration parameters are explained.

\subsection{Data acquisition system}
\begin{figure}
    \centering
    \includegraphics[width=1.0\linewidth]{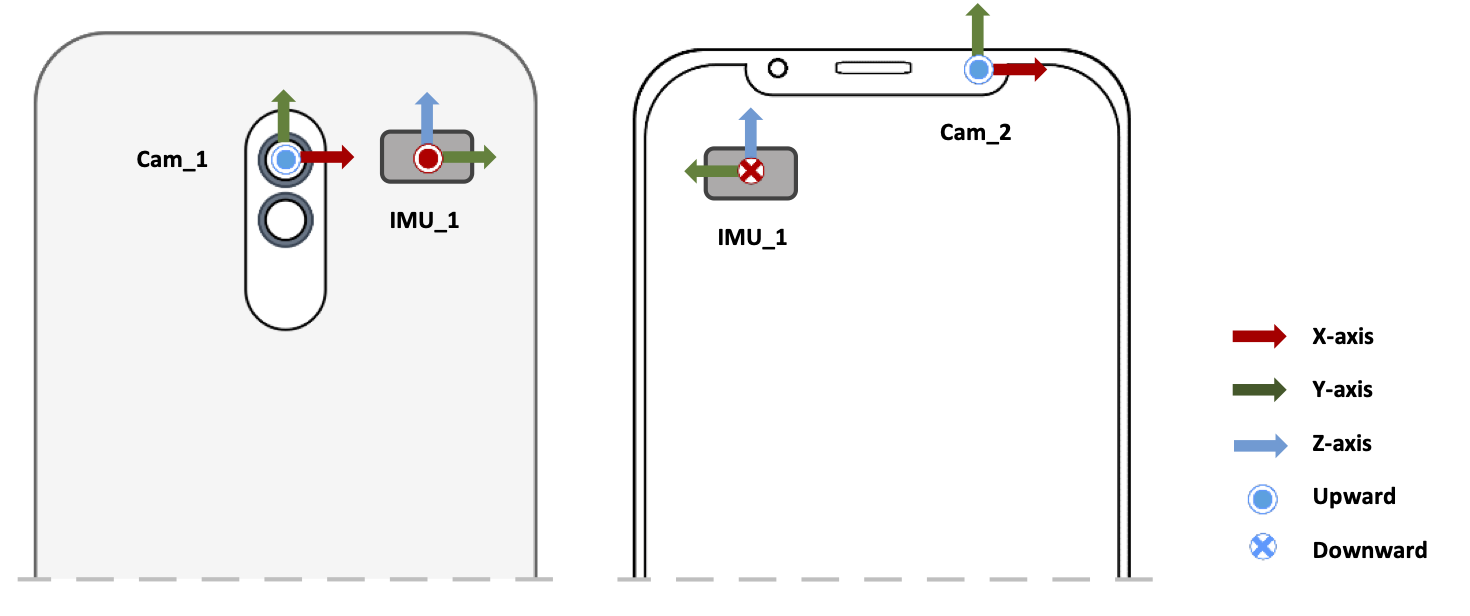}
    \caption{Details of cell-phone sensors used to record AUT-VI. The setup used to record this dataset has two cameras. The back camera shown at left is responsible for the main images of the dataset. The IMU is located near the cameras, and it is shown at both left and right figures.}
    \label{fig:sensors}
    \vskip -0.2in
\end{figure}

As mentioned earlier, the primary goal of AUT-VI is to help facilitate the navigation of visually impaired people and dataset acquisition. Hence, the dataset is recorded via a similar setup and the capture app is provided. To achieve this goal, we have designed a unique application to record the relevant sensors of the smartphone, including front and back cameras, IMU sensor, and GPS sensor. Fig.~\ref{fig:sensors} shows the smartphone used to record the dataset. The IMU sensor is BMI160, a low noise smartphone 16-bit IMU with a noise density of 180${\mu}g/\sqrt{Hz}$ for accelerometer and 0.007 $\deg/(s.\sqrt{Hz})$ for gyro. The IMU sampling frequency used to record this dataset is 200 Hz. The back camera providing the RGB images in the dataset is Sony IMX363 Exmor RS. This sensor is 12 megapixels with an aperture of f/1.9, pixel size of 1.4 ${\mu}m$, and sensor size of 1/2.55 inch. The data from the front camera is exclusively used for validating ground truth in outdoor settings. This camera features a 20-megapixel Samsung S5K3T1 sensor with an aperture of f/2.0. The GPS sensor data is recorded with an estimation error of fewer than 10 meters. 
The extrinsic and intrinsic calibrations are done using the well-known Agril grid and chess pattern utilizing the Kalibr tool~\cite{furgale2013unified} and careful validation. 
The ISO and time exposure parameters are not fixed and the settings are set to default inside the capture application.

Furthermore, the dataset is recorded via our proposed application (VIRec app). Within the VIRec application, users have the capability to define exposure time, ISO, and image size, and select specific camera sensors for operation. In instances where a device incorporates multiple camera sensors, simultaneous video recording from all cameras becomes feasible. VIRec facilitates the synchronized recording of the camera, Inertial Measurement Unit (IMU), and Global Positioning System (GPS) data utilizing a unified clock source on Android devices. The VIRec application is additionally accessible on the dataset's webpage.

Finally, the ground truth is entirely available for all sequences of the dataset. The ground truth is obtained from the GPS sensor. Hence, it has XYZ coordinates at 1Hz in only outdoor environments. 
Moreover, to ensure the accuracy of the ground-truth, it was inspected by human. In addition, it was also compared with the SLAM output of the front camera recording with no moving objects in the camera view. The sequences are all recorded with the same start and endpoint, hence their ground-truth inspection was easy.

\subsection{Data format}
The images inside the dataset are raw RGB from the back camera (main camera). The images are HD and portrait, with the resolution of 720$\times$1280, captured with a fixed frame rate of 30 FPS and encoding of MPEG-4.
Required data for intrinsic calibration is provided in the dataset. The camera ISO is variable inside the sequences and adapted to the environment (100 at noon, 3200 at dark). The exposure time varies depending on the ISO and camera limitations and fluctuates between 2 and 3 ms. Camera ISO and exposure times for each sequence are recorded in a file with a corresponding timestamp of each frame.

The IMU data provides accelerometer and gyroscope data. The file format is as follows:
$$
\left[ timestamp^{ns}, g_x^{rad/s}, g_y^{rad/s},g_z^{rad/s},a_x^{m/s^2},a_y^{m/s^2},a_z^{m/s^2} \right]
$$

where $g_{(.)}$ and $a_{(.)}$ correspond to rotation speed and acceleration, respectively. This data is provided at 200 Hz by the IMU sensor. The additive bias strength and noise density for calibration of SLAM systems are extracted from the sensor's datasheet and validated by the Kalibr tool. These values are also triple-checked with the output of VINS-Mono~\cite{qin2018vins}. These parameters are available on the dataset's wiki webpage.
The format of the ground-truth file is similar to the TUM VI~\cite{schubert2018tum} format.




\subsection{Sequence description and statistics}
\label{sec:dataset:sub:desc}
\begin{figure}
    \centering
    \includegraphics[width=\linewidth]{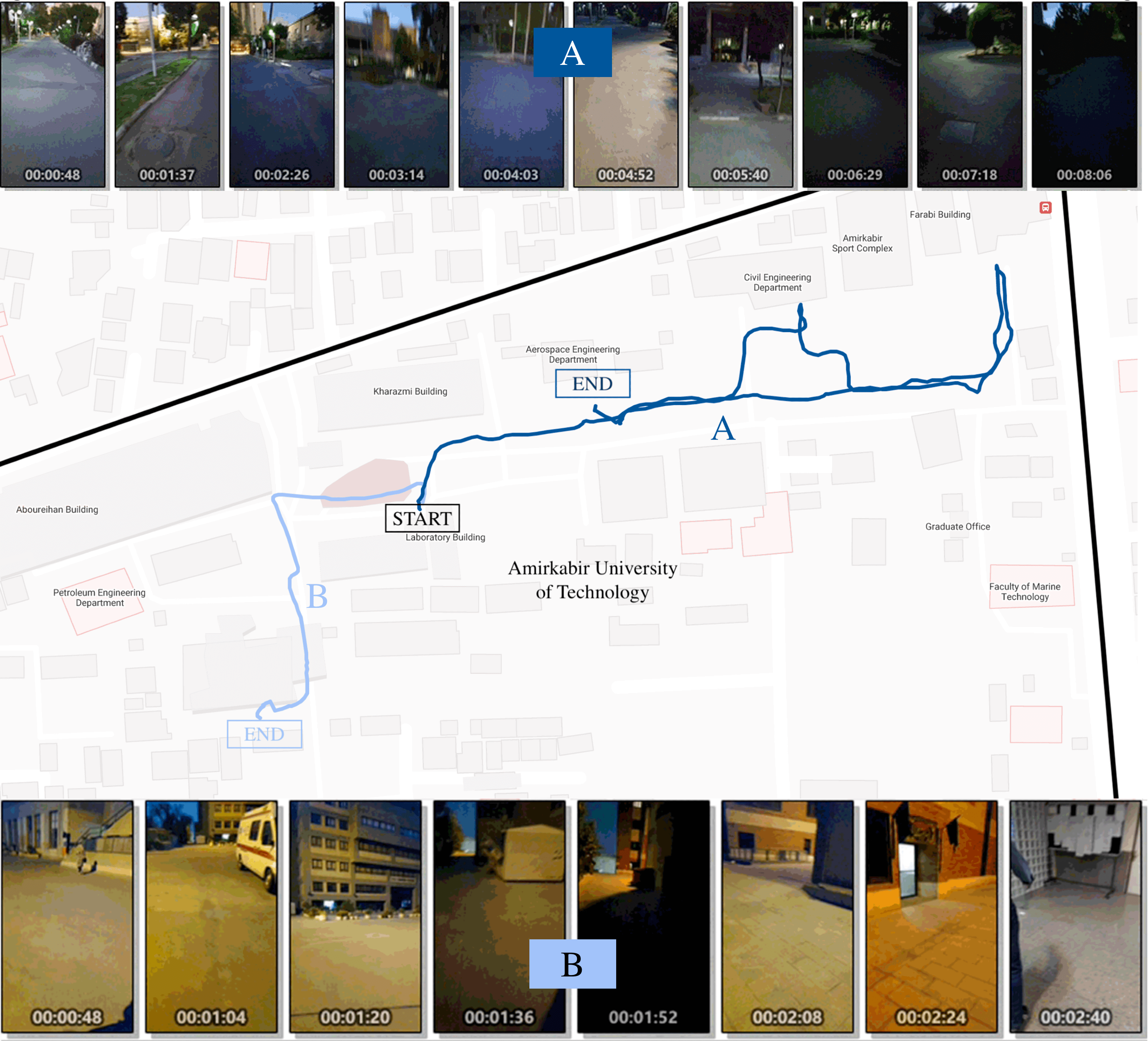}
    \caption{Two example of sequences available inside the dataset. A) A walk around the campus in the evening. B) A walk from center of campus to the bio-medical Engineering department.}
    \label{fig:examples}
    \vskip -0.1in
\end{figure}

\begin{figure} 
    \centering
    \includegraphics[width=\linewidth]{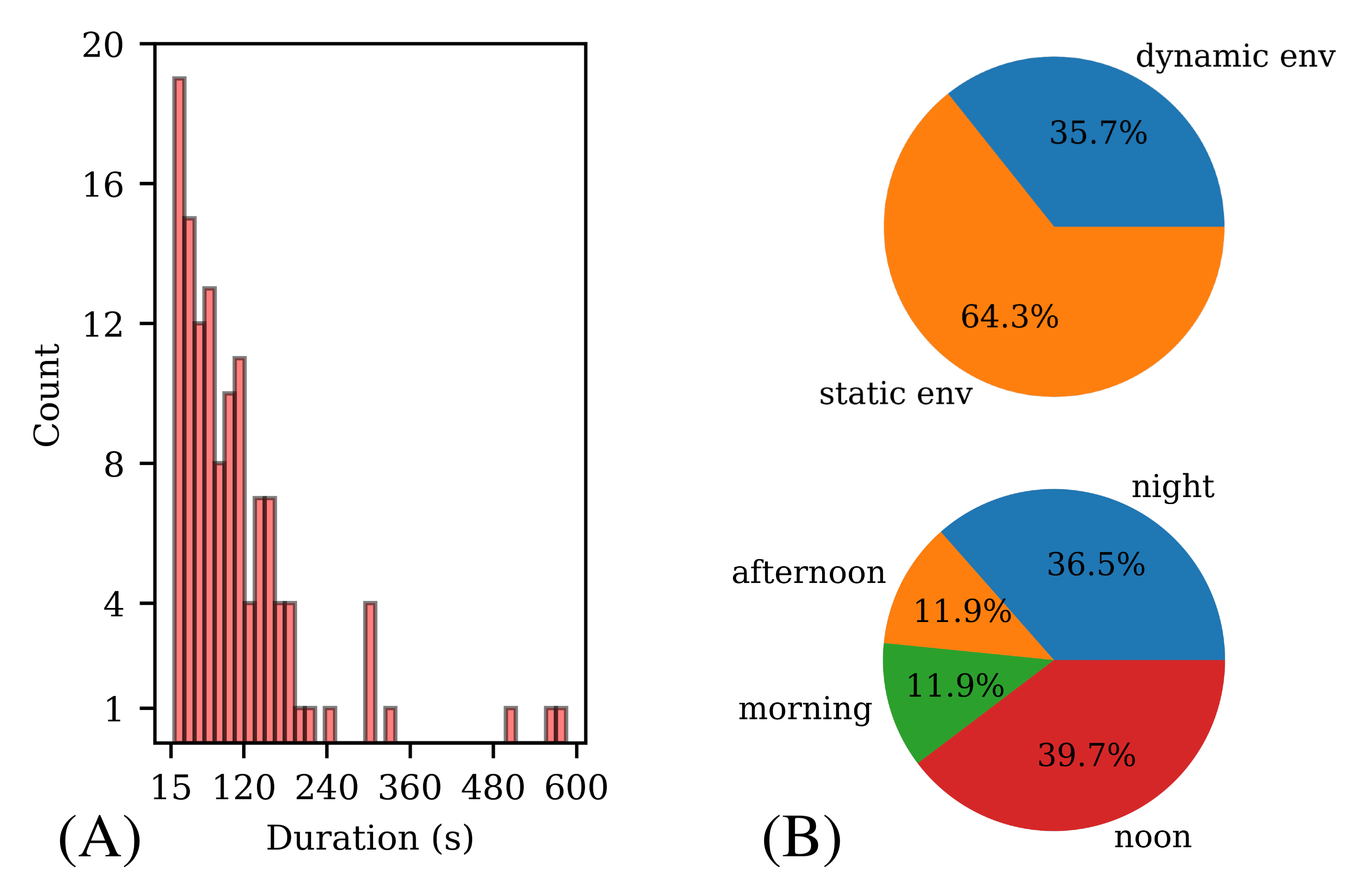}
    \caption{Dataset statistics. A) Histogram distribution of sequence lengths. B) Statistics about the exact environment and time of recording of sequences.}
    \label{fig:statistics}
    \vskip -0.1in
\end{figure}

Some example sequences of AUT-VI are shown in Fig.~\ref{fig:examples}. The length of the sequences varies from 20 seconds to more than 9 minutes with an average duration of 1 minute. Fig.~\ref{fig:statistics}-A shows a plot of the dataset sequence lengths. Each scenario has at least 2 sequences recorded in both night and day. For some scenarios, e.g., seq0 (university entrance), there are four sequences: morning, noon, evening, and night.
Moreover, each sequence is repeated in four different recording style. As a visually impaired person guided us, there are places where you cannot hold the phone in front of yourself since this draws suspicion. Therefore, the SLAM methods proposed for partially sighted people need to work even when a phone camera is held casually, i.e., near the leg with the camera facing to the side. Also, some disabled people may not be able to hold the camera steady while they are moving. So, we have recorded the videos with a shaking camera too. Hence, our dataset consists of four recording scenarios for each sequence; standard camera holding with static hands (walkstate1), standard camera holding with shaky hands (walkstate2), casual camera holding with static hands (walkstate3), and casual camera holding with shaky hands (walkstate4).

Furthermore, the sequences of the AUT-VI dataset are divided into two categories namely dynamic/challenging and normal. In the dynamic category there are multiple isolated dynamic and challenging sequences. The dynamic sequences are composed of 5 scenarios. In the first scenario, a human comes toward the camera and then stops. In the second scenario, the human goes away from the camera. In the third scenraio, the human blocks the camera view by his hand or body and the camera moves while being blocked. In the fourth scenario the camera moves behind a human. In the last scenario, human stays in front of the camera for a long time. Due to the standard of the TUM RGB-D~\cite{sturm2012benchmark} dataset of dynamic sequences, the mentioned scenarios are recorded with a static camera as well as different movements of the camera inclusive of 360-degree rotations and XYZ and spherical movements. We haven’t captured data for all possible permutations of the aforementioned challenges, but we have tried to consider each challenge isolated in some recorded scenario.

The outdoor sequences of our dataset contain explorations of the campus with various trajectories, including short and long tracks at different times of day and night. Therefore, the static outdoor sequences are perfect benchmarks for evaluating the performance of loop closure in SLAM methods. Moveover, there are three specific sequences to evaluate SLAM algorithms against reflection. 
In summary our dataset consists of 20 indoor, 81 outdoor, and 45 dynamic sequences. The statistics of duration, dynamicity, and time of recording the AUT-VI’s sequences are depicted in Fig.~\ref{fig:statistics}.

The calibration sequences for both camera and IMU are also included in the dataset. The camera calibration sequence is a simple XYZ and spherical trajectory over the Agril grid and chess pattern. The IMU calibration sequence is recorded while the device is in a fixed position for six hours. The IMU calibration data is recorded to determine the IMU noise pattern (using deep neural networks) or noise density and bias strength.

Finally, the recorded sequences contain very little clear faces and license plates. Due to the COVID-19 pandemic, people present in the videos are wearing masks; hence their faces are not recognizable. Also, we have legal permission to show the couple of specific faces in the dynamic category.
The unauthorized license plate numbers are covered using the plate detection algorithm provided in~\cite{samadzadeh2020rilp}. Moreover, we have personally inspected all of the frames inside the sequences to ensure there is no private data leak. The recording of the buildings and the personnel are authorized by the owner and Amirkabir University of Technology’s copyright department.
The license of the dataset is CC BY 4.0, and the application is released under the MIT license.


\section{Evaluation of SLAM frameworks on \mbox{AUT-VI}}
\label{sec:eval}
We selected four SOTA algorithms to represent three main groups of SLAM systems. For the VO frameworks, we used Basalt~\cite{usenko2019visual}. The VIO frameworks are represented using two SOTA algorithms, namely VINS-Mono~\cite{qin2018vins} and ORB-SLAM3~\cite{campos2021orb}. Finally, we utilized SLAMANTIC~\cite{schorghuber2019slamantic} for dynamic SLAMs group. The third group uses visual input and semantic segmentation to filter out the dynamic objects and estimate the trajectory.


\setlength{\tabcolsep}{5pt}
\begin{table}[t] 
    \centering
    \begin{small}
        \begin{tabular}{lccccc} 
            \hline
            Model & seq1 & seq2 & seq3 & seq4 & seq5\\
            \hline
            VINS-Mono \cite{qin2018vins}                &  0.010 & D & 0.017 & D & No\\
            ORB-SLAM3 \cite{campos2021orb}              &  0.008 & 0.072 & 0.012 & D & No\\
            Basalt \cite{usenko2019visual}              &  0.019 & 0.215 & 0.024 & 0.840 & N/A \\
            SLAMANTIC \cite{schorghuber2019slamantic}   &  0.012 & D & 0.027 & 0.166 & No\\
            \hline
        \end{tabular}
    \end{small}
    \caption{Performance of SOTA methods over selected sequences representing the dataset complexity. \textbf{D} stands for divergence.}
    \label{tab:res}
    \vskip -0.2in
\end{table}

The five sequences used in our experiments represent the complexity of the proposed AUT-VI dataset. The four sequences ``seq1-4'' correspond to
{\fontfamily{qcr}\selectfont entrance1-walkstate 1-noon}, {\fontfamily{qcr}\selectfont entrance1-walkstate2-noon}, {\fontfamily{qcr}\selectfont entrance 1-walkstate1-night}, and
{\fontfamily{qcr}\selectfont dynamic-outdoor-mov ingBehindPerson-noon} sequences in the dataset, respectively. The seq1 shows the university entrance at (1) noon, a sweep around it with (2) regular movements, and (3) no dynamic objects or complexities. The seq2 is same as the seq1 but with shaky camera movements. The seq3 contains regular (non-shaky and not tilted) camera movement with the same trajectory recorded at night. The seq4 is a long walk behind a person at noon. This sequence includes multiple frames in which a human covers the camera view. Lastly, there is seq5, in which we take seq1 and seq3 and merge them behind each other to see if the SLAM algorithms can detect the same architecture at the entrance and do the loop closure. To show the overall generality of these sequences we also evaluate the SOTA algorithms on all of the dataset shown in \ref{tab:res2}. Also, we report successful loop-closure rate and average error on the sequences that the algorithms did not diverge. The evaluation metric is Absolute Trajectory Error (ATE)~\cite{zhang2018tutorial}.

The experiments were conducted on a single PC with an Nvidia GTX 1080 Ti, a GTX 1080 GPU, and an i7-6500 CPU with 24GB of RAM and 512GB Samsung 850 SSD storage.


\subsection{Results}
The results of the experiments are shown in \ref{tab:res}. As can be seen, there are limitations in existing both dynamic and non-dynamic robust VI/VIO frameworks. The results show that all of the algorithms performed well on seq1 and seq3, as expected. These results indicate that if the camera's exposure time is set to a reasonable amount and camera blur at night is not very severe, the SOTA algorithms perform very well. Also, when there is no sharp brightness changes at night the SLAM frameworks can perform comparable to the day conditions. The VINS-Mono algorithm and SLAMANTIC are sensitive to camera shake and diverges in no time and reach the state of failure. The Basalt performs fair and ORB-SLAM3 outperforms all in seq3. The seq4 is the most difficult one. Therefore, the VIO frameworks diverge and the VO frameworks show enormous divergence. However, the SLAMANTIC method shows some promising results on seq4, but it chooses the dynamic points when the person covers the whole camera view, resulting in wrong trajectory estimations. Lastly, the loop-closure sequence examins the ability of each algorithm to perform loop-closure at conditions where (1) the scene is traversed in the day and (2) it is seen at night for the first time. None of the frameworks were able to perform stable successful day/night loop closure. The results at \ref{tab:res2} show that there were very few successful loop closures. SLAMANTIC is not performing better than other SOTAs. The reason behind it could be because most of the loop-closure scenes were non-dynamic. The average ATE of successfully tracked sequences shows SLAMANTICS success in estimating almost 70\% of the trajectory correctly.
For more detailed results and benchmarks, refer to the benchmark section of the AUT-VI webpage.

\setlength{\tabcolsep}{7pt}
\begin{table}[t] 
    \centering
    \begin{small}
        \begin{tabular}{lccc} 
            \hline
            Model & loop-closures & Avg ATE & Non D\\
            \hline
            VINS-Mono \cite{qin2018vins}                 & 31\% & 0.235 & 31\%\\
            ORB-SLAM3 \cite{campos2021orb}               & 33\% & 0.126 & 47\%\\
            Basalt \cite{usenko2019visual}               & N/A & 0.381 & 92\%\\
            SLAMANTIC \cite{schorghuber2019slamantic}    & 29\% & 0.096 & 70\%\\
            \hline
        \end{tabular}
    \end{small}
    \caption{Performance of SOTA methods evaluation on selected sequences representing the datasets complexity. Average ATE is calculated for only successfully tracked sequences. the \textbf{loop-closures} column corresponds to the rate of successful loop-closures inside the dataset. \textbf{Non D} corresponds to the rate of not diverged and successfully estimated trajectories.}
    \label{tab:res2}
    \vskip -0.2in
\end{table}

\section{Conclusion}

\label{sec:conclusion}
The AUT-VI dataset provides the challenges of dynamic environments, camera shake, loop-closure at different times of the day, and lighting conditions required for developing robust real-world SLAM algorithms.
Moreover, to facilitate future research on VIO algorithms, we have designed a unique Android application for researchers to create their own datasets.
For future works, we intend to support more diverse sequences. For this purpose, we plan to create sequences while we are driving and also, in more crowded environments. 
On the other hand, we want to adapt the dataset to disabled people's navigation systems; therefore, we plan to attach the camera to a wheelchair and generate sequences. Furthermore, we may include the segmentation of dynamic objects, to make the comparison of dynamic SLAM systems fair.
The SLAM algorithms can be improved by using inertial-only trajectory estimation algorithms in extremely dynamic or other challenging scenarios. Also, employing new correspondency checking algorithms, e.g., Superglue~\cite{sarlin2020superglue}, can improve the performance of the loop-closure mechanisms.

\addtolength{\textheight}{-12cm}   









{\small
\bibliographystyle{ieee_fullname}
\bibliography{egbib}
}

\end{document}